\documentclass[10pt,twocolumn,letterpaper]{article}

\usepackage[pagenumbers]{cvpr} 

\usepackage{times}  
\usepackage{helvet}  
\usepackage{courier}  
\usepackage[hyphens]{url}  
\usepackage{graphicx} 
\usepackage{multirow} 
\usepackage{multicol}
\usepackage{colortbl} 
\usepackage{rotating}
\usepackage{caption}
\usepackage{booktabs}
\usepackage{amsfonts}       
\usepackage{nicefrac}       
\usepackage{amsmath}
\usepackage{tabularx}
\usepackage{tcolorbox}

\usepackage{algorithm}
\usepackage{algorithmicx}
\usepackage{pifont}
\usepackage{amssymb}
\usepackage{tablefootnote}
\usepackage{float}
\usepackage{newfloat}
\usepackage{listings}
\usepackage{makecell}
 \usepackage[misc]{ifsym}
 \usepackage{cutwin}
\definecolor{cvprblue}{rgb}{0.21,0.49,0.74}
\usepackage[utf8]{inputenc} 
\usepackage[T1]{fontenc}    
\usepackage{amssymb}
\usepackage{fontawesome}
\DeclareUnicodeCharacter{2713}{\checkmark}  
\DeclareUnicodeCharacter{2717}{\ding{55}}   
\usepackage[pagebackref,breaklinks,colorlinks,allcolors=cvprblue]{hyperref}
\title{
DocSeeker: Structured Visual Reasoning with Evidence Grounding for Long Document Understanding
}

\author{  Hao Yan$^{1}$, Yuliang Liu$^{1\ast}$, Xingchen Liu$^{1}$, Yuyi Zhang$^{1}$,Minghui Liao$^{2}$, Jihao Wu$^{2}$, Wei Chen$^{1}$\thanks{Corresponding author.}, Xiang Bai$^{1}$\\
  $^{1}$Huazhong University of Science and Technology \quad $^{2}$Huawei Inc.\\
  \texttt{\{haoyan, ylliu, lemuria\_chen, xbai\}@hust.edu.cn}\\
  \url{https://github.com/yh-hust/DocSeeker}\\
}

\begin{document}
\maketitle
\begin{abstract}
Existing Multimodal Large Language Models (MLLMs) suffer from significant performance degradation on the long document understanding task as document length increases. This stems from two fundamental challenges: 1) a low Signal-to-Noise Ratio (SNR), with crucial evidence buried in irrelevant pages; and 2) supervision scarcity, as datasets offering only final short answers provide a weak learning signal. In this paper, we address these challenges by proposing a paradigm that requires the model to execute a structured ``\textbf{Analysis}, \textbf{Localization} and \textbf{Reasoning}'' workflow. To instill this capability, we design a two-stage training framework: we first perform Supervised Fine-Tuning on high-quality data generated via an efficient knowledge distillation strategy. Subsequently, we employ an Evidence-aware Group Relative Policy Optimization which jointly optimizes for both evidence localization and answer accuracy. Additionally, we introduce a Evidence-Guided Resolution Allocation strategy to mitigate memory constraints of training on multi-pages documents. Extensive experiments demonstrate that DocSeeker achieves superior performance on both in-domain and out-of-domain tasks. We show it robustly generalizes from short-page training to ultra-long documents and is naturally synergistic with visual Retrieval-Augmented Generation systems, serving as a solid foundation for their implementation.
\end{abstract}

\begin{figure*}[htbp]
  \centering
  \includegraphics[width=0.92\textwidth]{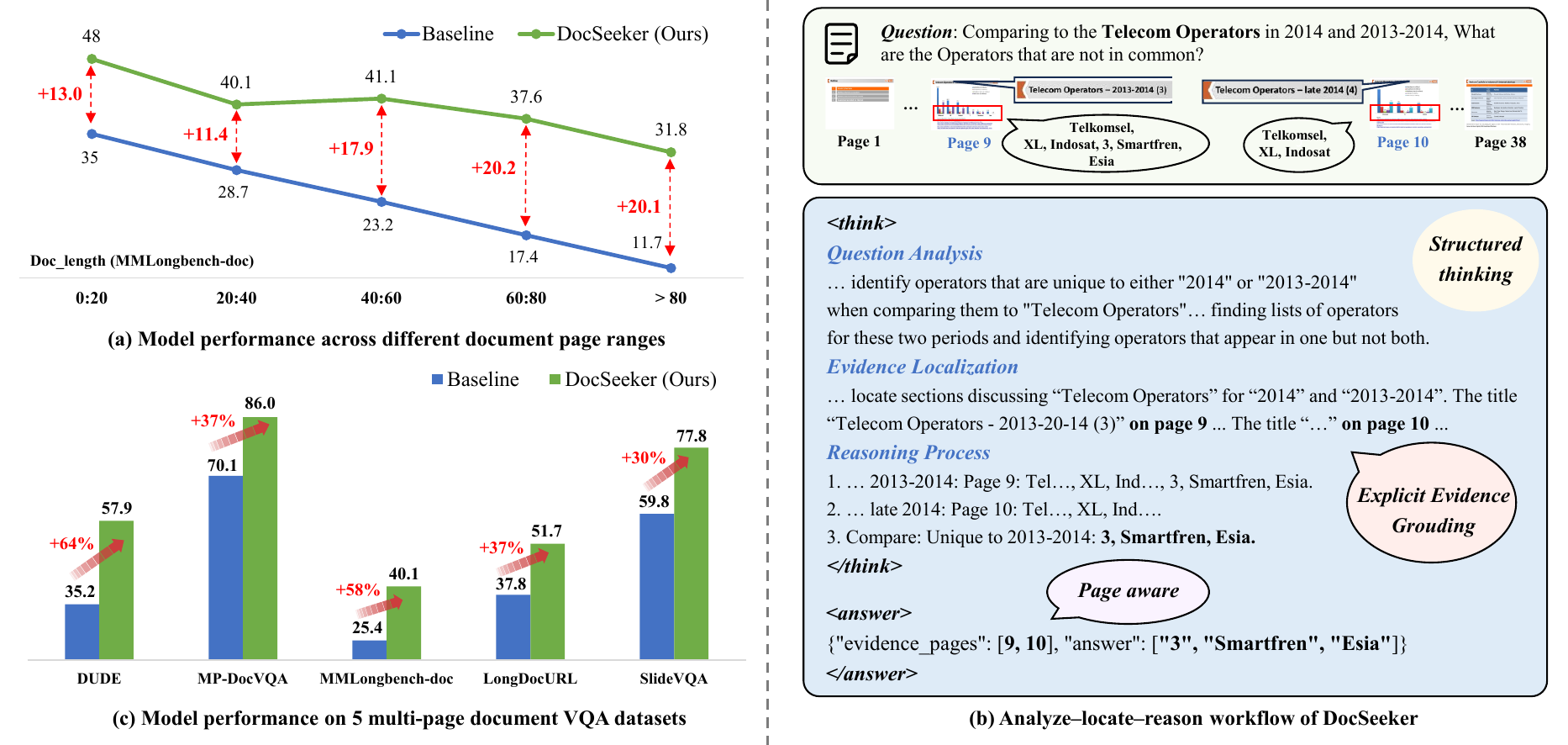}
  \caption{Overview of Principal Experimental Results in Long Document Understanding (Left) and the ALR Reasoning Paradigm of DocSeeker (Right).}
  \label{fig:pipeline}
\end{figure*}

\section{Introduction}
Multi-Page Document Visual Question Answering presents a grand challenge in reasoning over lengthy, visually-rich documents~\cite{tito2023hierarchical,van2023document,ke2025large}. Traditional parsing-based pipelines often rely on Optical Character Recognition (OCR) models~\cite{wang2024mineru,feng2025dolphin,li2025monkeyocr}, which is prone to \emph{cascading errors} and risks \emph{losing layout and stylistic information}. In contrast, building upon the huge success of Large Language Models~\cite{liu2023brief}, a more elegant and unified direction involves \textbf{pure-visual solutions} powered by Multimodal Large Language Models (MLLMs)~\cite{bai2025qwen25,zhu2025internvl3,hu2025mplug2, kim2022ocr,vteam2025glm45vglm41vthinkingversatilemultimodal,wu2024deepseekvl2mixtureofexpertsvisionlanguagemodels,liu2024textmonkey, lee2023pix2struct, yao2024minicpm}. By treating each page as an image, these models preserve the document's visual integrity holistically, offering a general framework that bypasses fragile intermediate steps and applies universally to almost any document format.

However, MLLMs powered pure-visual methods confront two fundamental hurdles when scaling to long documents. The first is severe \textbf{low Signal-to-Noise Ratio (SNR)}, where crucial evidence are buried within vast irrelevant content. Although several visual-Retrieval-Augmented Generation (RAG) methods~\cite{yu2024visrag,faysse2024colpali,zheng2026doc} for document page retrieval have been proposed, which can pre-filter question relevant pages, it introduces the classic top-$k$ dilemma: a large $k$ ensures high recall but introduce more noise, whereas a small $k$ risks missing the evidence. 

The second is the \textbf{scarcity of fine-grained supervision}. Most existing multi-page document VQA datasets provide only final short answers, lacking intermediate reasoning steps like evidence localization and information synthesis. Training models with such labels may cause them to learn fragile shortcuts through memorization, rather than developing genuine reasoning capabilities. This can lead to two weaknesses: First, it creates models that are effectively black boxes~\cite{ye2023mplugdoc,xie2024wukong, yu2025docthinker}, severely undermining their interpretability, whose outputs lack clear reasoning path and source attribution. Second, this learning strategy can result in poor generalization performance when dealing with out-of-distribution (OOD) documents.

To address above limitations of existing methods, we introduce DocSeeker, a document multimodal large language model built upon Qwen-2.5-VL-7B~\cite{bai2025qwen25}. Inspired by human cognitive processes, DocSeeker adopts a novel \emph{Analysis–Localization–Reasoning} (ALR) visual reasoning paradigm, as illustrated in Fig.\ref{fig:pipeline}(b). When processing a multi-page document and a user's question, each page's visual tokens are prefixed with a page id,  acting as a pointer, anchoring all information to its specific page. Instead of a direct answer, DocSeeker first generates a structured thinking process. It starts by analyzing the user's intent, then identifies supporting evidence with explicitly mentioning potential pages, and finally, builds a line of reasoning from that evidence. Following this thinking process, the model will deliver the final answer, along with the specific page ids where the evidence can be found.

To enable the model to effectively learn this reasoning paradigm, we propose a two-stage training framework. In the first stage, we employ an efficient data distillation strategy to generate high-quality ALR Chain-of-Thought (CoT) annotations from a powerful teacher model for supervised fine-tuning (SFT). In the second stage, we introduce \emph{Evidence-aware Group Relative Policy Optimization} (EviGRPO) that jointly optimizes evidence localization and reasoning abilities. To render this process computationally tractable and more effective for long visual sequences, we propose \emph{Evidence-Guided Resolution Allocation} (EGRA) strategy, which is applied during training to optimize resource allocation and strengthen the supervisory signal.

Extensive experiments demonstrate the effectiveness of DocSeeker. First, as shown in Fig.~\ref{fig:pipeline}(c), DocSeeker achieves a remarkable 30-60\% performance gain across all five document VQA benchmarks compared to the Baseline, which shares the same architecture and parameters. Furthermore, DocSeeker's performance not only surpasses other advanced open-source MLLMs but also proves highly competitive with closed-source commercial models. More importantly, despite being trained only on existing short-page documents, DocSeeker achieves robust generalization to ultra-long document reasoning, effectively mitigating the performance decay associated with long-sequence inputs, as shown in Fig.~\ref{fig:pipeline}(a). DocSeeker's strong localization capability, derived from the ALR paradigm, is proven to be naturally synergistic with visual RAG systems, serving as an ideal foundation for building next-generation, efficient, and robust implementations. 

Ours core contributions can be summarized as follows: 1) We introduce the \textbf{ALR} paradigm, a structured workflow inspired by human cognition, along with the corresponding ALR CoT dataset. This paradigm compells the model to internalize an ALR process rather than superficially imitating outputs; 2) We propose a two-stage training framework: \textbf{Stage \Rmnum{1} injecting the ALR Paradigm via SFT}, which uses an efficient, high-fidelity data distillation strategy to generate ALR CoT data for SFT; \textbf{Stage \Rmnum{2} EviGRPO}, which jointly optimizes for both evidence localization and reasoning abilities; and \textbf{EGRA} strategy mitigates memory constraints while simultaneously strengthening the supervision signal; 3) Extensive experiments demonstrate our model's strong localization capabilities and robust generalization in OOD ultra-long document scenarios, and confirm it is naturally synergistic with visual RAG systems.

\section{Related Work}

\subsection{Multi-Page DocVQA dataset}
Multi-page Document Visual Question Answering challenges models to comprehend and reason over lengthy, visually-rich documents. Early datasets~\cite{tito2023hierarchical,van2023document,tanaka2023slidevqa, li2024multimodalarxiv, ding2024mvqa} significantly advanced progress in multi-page document understanding. However, these works primarily focused on documents within a 20-page limit. More recent benchmarks, including MMLongBench-doc~\cite{ma2024mmlongbench}, LongDocURL~\cite{deng2025longdocurl}, MMdocIR~\cite{dong-etal-2025-mmdocir}, M-LongDoc~\cite{chia-etal-2025-longdoc} and DocBench~\cite{zou-etal-2025-docbench}, have extended the challenge to hundred-page scale scenarios. Despite this progress, annotating such ultra-long documents remains a significant bottleneck due to high human costs and the low accuracy of MLLM-based auto-labeling, making relevant, high-quality training data exceedingly scarce. Exacerbating this issue, most existing datasets with training splits primarily offer ``long-sequence input to short-answer output'' pairs, which provides only a sparse signal and forces models to implicitly learn evidence localization from the final answer.

\subsection{Long Document Understanding}
Existing methods for long document understanding can be primarily categorized into OCR-based methods and pure-vision methods. OCR-based methods, such as LongFormer~\cite{beltagy2020longformer} and LayTokenLLM~\cite{zhu2025simple}, rely on OCR tools to extract text and layout for downstream models. However, this pipeline suffers from domain brittleness, cascading errors, and heavyweight OCR-tool dependency. In contrast, purely visual approaches feed pages directly into the MLLM, this paradigm fundamentally avoids error propagation and holistically preserves critical visual and layout information. Nevertheless, the performance and efficiency of this pure-vision paradigm are significantly challenged when processing ultra-long documents, 
leading existing MLLMs to primarily employ two strategies: (1)~\textbf{Visual Retrieval-Augmented Generation} leverage an external visual retriever, such as ColPail~\cite{faysse2024colpali} and DSE~\cite{ma2024dse}, to identify the Top-K relevant pages and then process them for fine-grained reasoning. Classic works that adopt this paradigm include VisRAG~\cite{yu2024visrag}, SV-RAG~\cite{chen2024sv}, and VDocRAG~\cite{tanaka2025vdocrag}. (2)~\textbf{End-to-End methods.} These methods must either use high-performance visual token compressors, like mPLUG-DocOwl2~\cite{hu2025mplug2}, or significantly expand the context window, such as in Qwen2.5VL~\cite{bai2025qwen25} and InternVL3~\cite{zhu2025internvl3}. These two strategies collectively represent the dominant research paradigms in the pure-vision long document understanding. Details of those models are provided in Appendix E.

\section{DocSeeker}

\begin{figure*}[htbp]
  \centering
  \includegraphics[width=0.92\textwidth]{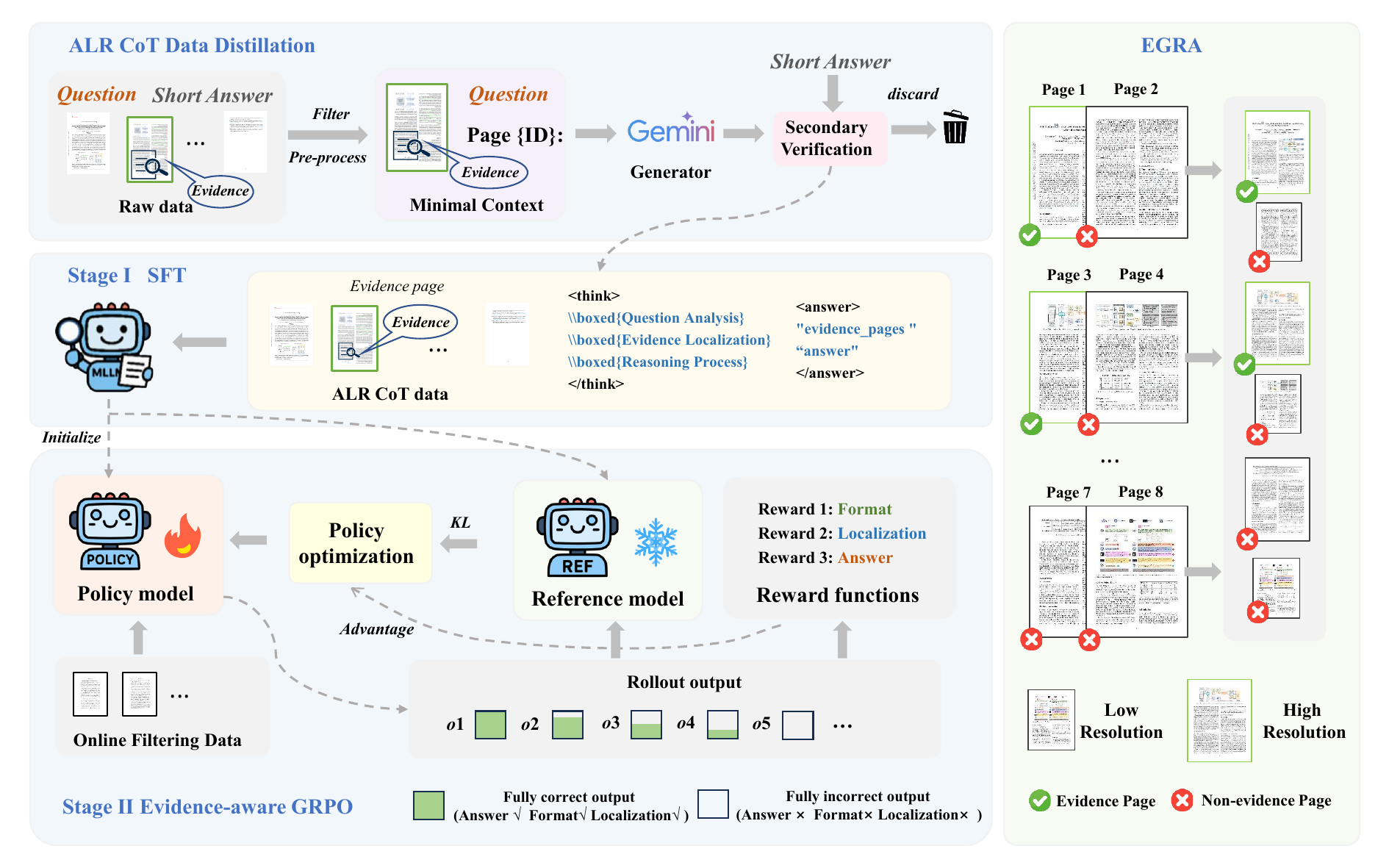}
  \caption{Overview of our training framework. (Left) Stage I SFT, where the model is supervised fine-tuned on high-quality ALR CoT data; Stage II RL, where the Policy model is optimized via reinforcement learning, guided by a multi-dimensional reward function that includes format, localization, and accuracy. (Right) Our proposed Evidence-Guided Resolution Allocation (EGRA) strategy.}
  \label{fig:method}
\end{figure*}

We build DocSeeker upon the Qwen-2.5-VL-7B-Instruct architecture~\cite{bai2025qwen25} as our backbone model, an powerful open source MLLM comprising a Transformer-decoder~\cite{vaswani2017attention} based LLM, a Vision Transformer (ViT)~\cite{dosovitskiy2020image} encoder, and a VL-Adapter for mapping image features into visual tokens. Rather than introducing any extraneous modules, we aim to enhance the backbone model's intrinsic ability to mitigate the signal degradation caused by the noise and redundancy inherent in long visual token sequences, a critical bottleneck for scaling to ultra-long document VQA. To this end, we propose the novel ALR visual reasoning paradigm. Inspired by the human cognitive process of \emph{find-then-reason}~\cite{stenning2012human}, the ALR paradigm enables the model to efficiently localize salient information before conducting detailed reasoning. We instill this paradigm through a two-stage training framework composed of \emph{SFT} and \emph{Evi-GRPO}, where \emph{EGRA} is utilized in both training stages to enable efficient long-sequence training.

\subsection{ALR Visual Reasoning Paradigm}\label{Sec:ALR}

The ALR visual reasoning paradigm is the core for our DocSeeker, it's implemented via two primary mechanisms: \emph{Page-Aware Input Representation} and \emph{Structured Reasoning Paradigm with Explicit Evidence Grounding}, as illustrated in the Fig.~\ref{fig:pipeline}(b).

\noindent\textbf{Page-Aware Input Representation.} \quad To enable the model to ground its reasoning on specific document pages, we introduce a page-aware input representation. This is achieved by interleaving textual page identifiers with their corresponding visual tokens. Specifically, for a user question $Q$ and a document of $N$ pages with each page image denoted as $\mathbf{I}_i$ ($i \in {1, \dots, N}$), the input sequence $\mathbf{X}$ fed to the LLM is constructed as:
\begin{equation}
\mathbf{X} = \mathbf{e}_Q \oplus \bigoplus_{i=1}^N \left( \mathbf{e}_i \oplus \mathbf{v}_i \right)
\label{eq:input_representation}
\end{equation}
Here, $\mathbf{e}_Q$ represents the text embeddings of the question $Q$. For each page $i$, $\mathbf{v}_i$ denotes its visual tokens, which are preceded by the text embeddings of a page identifier, $\mathbf{e}_i$ (e.g., (e.g., \emph{``Page $i$''}). The visual tokens are generated by the vision backbone: $\mathbf{v}_i = f_{\text{adapter}}(f_{\text{enc}}(\mathbf{I}_i; r_i))$, where $r_i$ is the image resolution. The $\oplus$ operator denotes sequence concatenation. 

\noindent\textbf{Structured Reasoning Paradigm with Explicit Evidence Grounding.} \quad Inspired by recent large reasoning models (LRMs)~\cite{jaech2024openai,guo2025deepseek} that separate the thought process from the answer (e.g., using \texttt{<think>} and \texttt{<answer>} tags), we enforce a novel structured reasoning paradigm tailored for multi-page long document VQA. Instead of generating free-text thought chains, our model's output, denoted as $\mathbf{Y}$, must adhere to a specific, interpretable reasoning path. The output sequence is defined as: 
\begin{equation}
\mathbf{Y} = \mathbf{Y}_{\text{th}} \oplus \mathbf{Y}_{\text{ans}} = \left( \mathbf{Y}_{\text{A}} \oplus \mathbf{Y}_{\text{L}} \oplus \mathbf{Y}_{\text{R}} \right) \oplus \left( \mathbf{Y}_{\text{E}} \oplus \mathbf{Y}_{\text{F}} \right)
\label{eq:output_representation}
\end{equation}
Here, the thought process $\mathbf{Y}_{\text{th}}$ is decomposed into three parts: 1) First, the model generates \textit{Question Analysis} $\mathbf{Y}{_\text{A}}$ to deconstruct the user's intent; 2) The critical evidence-finding part is denoted as \textit{Evidence Localization} $\mathbf{Y}_{\text{L}}$, where the model is required to scan the document and explicitly report which pages it deems relevant and why, forcing the model learns visual evidence grounding by referencing specific page identifiers; 3) Finally, it articulates a \textit{Reasoning Process} $\mathbf{Y}_{\text{R}}$ that synthesizes information from the located evidence to derive a conclusion. Following the thought process, the model generates the answer $\mathbf{Y}_{\text{ans}}$, which consists of a list of the page numbers identified $\mathbf{Y}_{\text{E}}$ as evidence, and the final, concise answer $\mathbf{Y}_{\text{F}}$. 

The proposed ALR Visual Reasoning paradigm, a \textbf{structured} and \textbf{page-aware} paradigm, offers significant advantages. It mandates explicit evidence grounding, which yields \textbf{high interpretability} and allows users to easily verify the answer by referring to the cited pages. More critically, subsequent experiments demonstrate that by compelling the model to localize and cite page-specific information, it learns to effectively distinguish visual tokens belonging to different pages, thus counteracting the interference and noise in long visual inputs. Furthermore, and to our surprise, \textbf{the model can seamlessly integrate with a visual retrieval system}. It can still perform reasoning effectively even with the discontinuous, non-sequential Top-$k$ pages returned by the visual page retriever.

\subsection{Training Procedure}
To enable the model to master the ALR paradigm mentioned in Sec.~\ref{Sec:ALR}, we designed the following training procedure, as shown in Fig.~\ref{fig:method}.

\noindent\textbf{Stage \Rmnum{1}: Injecting the ALR Paradigm via SFT.} \quad This stage aims to address supervision scarcity and instill the core capabilities for backbone model to follow the ALR visual reasoning paradigm by constructing high quality training data. Instead of building from scratch, we build upon existing multi-page VQA datasets that provide labels consisting of final short answer and evidence page IDs, but lacking the explicit, step-by-step reasoning paths required for the ALR format. To achieve this, we construct a high-quality training dataset via knowledge distillation. We utilize Gemini-2.5-Flash~\cite{comanici2025gemini} as our teacher model. Instead of a naive and costly full-document prompting approach, we employ a more efficient strategy: we provide the teacher with a \textbf{minimal context} composed of only ground-truth evidence pages, a very few distractor pages within the same document, corresponding page IDs and the question. The teacher model is required to generate an complete ALR-formatted response in the instruction prompt. To guarantee data quality, we implement a two-stage verification pipeline: an initial automated Exact Match (EM) check on the ground-truth short answer and evidence page ids, followed by a semantic validation for failed samples with wrong matched answers using GPT-4o~\cite{hurst2024gpt}. This secondary check effectively rescues correct examples where the generated output is semantically correct but fails the strict EM criteria due to trivial formatting differences or acceptable paraphrasing. See more details in Appendix A. 

With this high-quality, verified dataset, we then \textbf{fine-tune} our backbone model, using a standard cross-entropy loss to predict the entire ALR-formatted response $\mathbf{Y}$, given the \textbf{complete document} (i.e., all pages and page ids) and question as input. By supervising on the full reasoning process over complete documents, this stage enhances reasoning rather than memorizing by compelling it to learn the holistic ALR reasoning structure.

\noindent\textbf{Stage \Rmnum{2}: Evidence-aware GRPO.} \quad While Supervised Fine-Tuning (SFT) equips the model with the ability to follow the ALR paradigm, the resulting reasoning paths from this imitation learning are often sub-optimal. To transcend this limitation, we introduce a second fine-tuning stage using Group Relative Policy Optimization (GRPO)~\cite{guo2025deepseek}, a reinforcement learning approach enables the model to learn directly from outcome signals. Specifically, GRPO samples multiple candidate responses, evaluates them with a reward function, and updates the model's policy to favor high-reward outputs. To tailor this process to our ALR paradigm, we design a bespoke reward function beyond standard GRPO. Our EviGRPO reward jointly optimizes for format correctness, evidence grounding, and answer accuracy. The total reward, $R = \lambda_1 R_{\text{format}} + \lambda_2 R_{\text{evidence}} + \lambda_3 R_{\text{answer}}$, is a weighted sum of three rewards. First, a binary Format Reward $R_{\text{format}}$ ensures the model's output strictly adheres to the ALR template. Second, we introduce an evidence grounding Reward $R_{\text{evidence}}$ to directly score the accuracy of the identified evidence pages. This is calculated using a weighted ($\beta > 1$) F1 score that prioritizes recall over precision. Finally, a standard Answer Reward measures the correctness of the final answer ($\mathbf{Y}_{\text{F}}$) using the Average Normalized Levenshtein Similarity (ANLS)~\cite{yujian2007normalized} metric. This multi-faceted reward signal guides the model to learn both localization and reasoning within the ALR paradigm in a more direct and efficient manner. 

\noindent\textbf{Evidence-Guided Resolution Allocation.} \quad Training on long documents presents a significant challenge: processing full-resolution pages generates vast input visual tokens, resulting prohibitive out of memory on GPUs during training. To address this, we propose Evidence-Guided Resolution Allocation (EGRA), a simple yet effective training strategy employed during both the SFT and GRPO stages to support longer document inputs. The core idea is to apply a differentiated image resolution scheme: ground-truth evidence pages are maintained at a high resolution to preserve critical details, while for non-evidence pages, we randomly downsample a large portion (70\%) to a lower resolution ($r_i$ from 1024 to 256), with the remaining ones (30\%) retaining their original high resolution ($r_i = 1024$). This approach significantly reduces the total input token count, enabling us to use longer documents for training. During inference, all pages are processed at their native high resolution, which is computationally feasible since inference has a lower memory footprint. Beyond mitigating GPU memory constraints, our experiments find this strategy can facilitate learning by increasing the signal-to-noise ratio in the training data,  outperforming naive strategies such as fix resolution or directly removing a subset of non-evidence pages.

\section{Experiment}

\subsection{Experimental Setup}

\noindent\textbf{Datasets.} Our raw training data is sourced from  MP-DocVQA~\cite{tito2023hierarchical} and DUDE~\cite{van2023document} (both up to 20 pages). The evaluation is divided into two parts: (1) In-Domain evaluation, conducted on MP-DocVQA and DUDE; and (2) OOD evaluation on challenging benchmarks: MMLongBench-Doc~\cite{ma2024mmlongbench} (up to 468 pages), LongDocURL~\cite{deng2025longdocurl} (avg. 30 pages), and SlideVQA~\cite{tanaka2023slidevqa} (up to 20 pages).

\noindent\textbf{Data Distillation and Filtering.} Our SFT data preparation begins with pre-filtering to remove overly short documents, which yields a curated set of 19,386 long-document VQA samples. Subsequently, we employ Gemini-2.5-flash~\cite{comanici2025gemini} as the teacher model to distill 13,986 high-quality ALR-paradigm samples (8,676 from MP-DocVQA and 5,310 from DUDE) for the SFT stage. For the RL stage, we source data from the remaining training instances of MP-DocVQA and DUDE. To select valuable samples, we employ an \textit{online filtering} strategy, following~\cite{meng2025mm}, to filter out instances yield ``zero advantage'' for the policy update.

\noindent\textbf{Training and Evaluation Configuration.} All training is conducted on 16 NVIDIA A800 GPUs. We adopt Qwen2.5-VL-7B-Instruct~\cite{bai2025qwen25} as our baseline. In the SFT stage, we train the model on our distilled data for 2 epochs with a learning rate of $1\times10^{-6}$. In the RL stage, we train for 6 epochs with a learning rate of $1\times10^{-6}$ and a rollout group size of 16. The reward weights for format, evidence, and answer were set to 0.1, 0.3, and 0.6, respectively, and the $\beta$ for the weighted F1-score (evidence reward) was set to 2.0. For evaluation, we process each document page at a resolution of $1024\times784$ to balance visual clarity with the capacity to accommodate longer document contexts.

\begin{table*}[htbp]
\centering
\caption{Comparison of different methods on five document understanding benchmarks. The results are reported on DUDE (ANLS), MPDocVQA (ANLS), MMLongBench-doc (Acc), LongDocURL (Acc), and SlideVQA (F1). “Param.” denotes parameter scale; Baseline-SFT (short-answer) refers to the baseline model fine-tuned on raw short-answer data; “Evi.” indicates evidence type (T = text, P = pages, I = images extracted from pages); “RAG” indicates whether retrieval is used; and “MMLong.“ and “LongDoc.“ are abbreviations for MMLongBench-doc and LongDocURL. The best and second-best results are highlighted in bold and underlined, respectively.}
\label{tab:sp_results}
\fontsize{9}{11}\selectfont
\setlength\arrayrulewidth{0.6pt}
\resizebox{1\textwidth}{!}{
\begin{tabular}{l l c c c c c c c c}
\hline
Method & Venue & {Param.} & Evi. & RAG &  \textbf{DUDE} &  \textbf{MPDocVQA} &  \textbf{MMLong.} & \textbf{LongDoc.} & \textbf{SlideVQA} \\
\hline
\textcolor{gray}{\fontsize{7}{8}\textit{Open Source}} \\
HiVT5~\cite{tito2023hierarchical} & PR & 0.3B & T+P & × & 23.1 & 62.0 & - & - & -  \\
CREAM~\cite{zhang2024cream} & ACM MM 2024 & 7B & T+P & \checkmark & 52.5 & 74.3 & - & - & -  \\
mPLUG-DocOwl2~\cite{hu2025mplug2} & ACL 2025 &  8B & P & × & 46.8 & 69.4 & 13.4  &  5.3 & - \\
M3DocRAG~\cite{cho2024m3docrag}  & Arxiv 2024 & 9B & P & $\checkmark$ & - & 84.4 & 21.0 & 35.1 & 55.7  \\
Vis-RAG~\cite{yu2024visrag} & ICLR 2025 & 4B & P & $\checkmark$ &  - & 70.9  & 18.8  & 41.9  & 50.7    \\
SV-RAG~\cite{chen2024sv} & ICLR 2025 & 4B & P & $\checkmark$ & 45.0 & 71.0 & 23.0 & - & -  \\
VDocRAG~\cite{tanaka2025vdocrag} & CVPR 2025 & 8B & P & $\checkmark$ & 44.0 & 62.6 & 18.4 & 39.8 & 42.0  \\
Docopilot~\cite{mathur2024docpilot} & CVPR 2025 & 8B & T+I & × & - & 81.3 & 28.8 & - & -  \\
DocVLM~\cite{nacson2025docvlm} & CVPR 2025 & 8B & T+P & × & 47.4 & \underline{84.5} & - & - & -  \\
InternVL3~\cite{zhu2025internvl3} & Arxiv 2025 & 8B & P & × & 47.4 & 80.8 & 24.1 & 38.7 & 54.4 \\
\textcolor{gray}{\fontsize{7}{8}\textit{Closed-source commercial models}} \\
Qwen-VL-Max~\cite{Qwen-VL} & - & - & P & × & - & - & - & 49.5 & -  \\
GPT-4V & - & - & P & × & - & - & 32.4 &  - & - \\
Gemini-1.5-Pro~\cite{team2024gemini} & - & - & P & × & 46.0 & - & 28.2 & 50.9 & -  \\
GPT-4o~\cite{hurst2024gpt} & - & - & P & × & 54.1 &  67.4 & \textbf{42.8} & \textbf{64.5} & -  \\
\hline
\rowcolor{LLightGray} \multicolumn{10}{c}{\textit{Ours}} \\
\hline
Baseline & - & 7B & P & × & 35.2 & 70.1 & 25.4 & 37.8 & 59.8 \\
\makecell[l]{Baseline-SFT (short-answer)}
  & - & 7B & P & × & 56.0 & 82.9 & 28.8 & 42.7 & 67.4 \\
\rowcolor{rowgray}
DocSeeker-SFT    & - & 7B & P & × & \underline{56.8} & 82.1 & 38.6 & 49.1 & \underline{75.2} \\
\rowcolor{rowgray}
DocSeeker & - & 7B & P & × &  \textbf{57.4} & \textbf{86.2} & \underline{40.1} & \underline{51.7} & \textbf{77.1} \\
\hline \hline 
\end{tabular}
}
\end{table*}

\subsection{Main Results}

Tab.~\ref{tab:sp_results} presents a comprehensive comparison of DocSeeker against state-of-the-art methods across five document understanding benchmarks. We compare with three categories of approaches: (1) End-to-end MLLMs that process full document pages as input, including InternVL3~\cite{zhu2025internvl3} and mPLUG-DocOwl2~\cite{hu2025mplug2}; (2) Parsing-based methods that rely on OCR or PDF parsers to assist the visual language model, such as HiVT5~\cite{tito2023hierarchical}, CREAM~\cite{zhang2024cream}, Docpilot~\cite{mathur2024docpilot}, and DocVLM~\cite{nacson2025docvlm}; Retrieval-augmented methods, including M3DocRAG~\cite{cho2024m3docrag}, Vis-RAG~\cite{yu2024visrag}, SV-RAG~\cite{chen2024sv}, and VDocRAG~\cite{tanaka2025vdocrag}. We also include closed-source commercial models such as GPT-4o, GPT-4V, Qwen-VL-Max, and Gemini for reference. Detailed evaluation settings for all compared models are presented in Appendix E. We can draw the following key conclusions:

\noindent\textbf{Performance and Generalization of DocSeeker.} In In-Domain evaluation, DocSeeker surpasses all existing open-source and commercial models on MPDocVQA and DUDE. More importantly, in OOD evaluation, DocSeeker establishes itself as the open-source state-of-the-art across all OOD benchmarks and proves highly competitive with advanced closed-source commercial models.

\noindent\textbf{Limitations of SFT on Short-Answer Data.} SFT  on short-answer data boosts In-Domain performance but provides only marginal OOD generalization, suggesting the model is primarily memorizing answers rather than developing true long-document understanding capabilities.

\noindent\textbf{The role of the ALR reasoning paradigm.} SFT on our high-quality, distilled data equips the model with the ALR reasoning paradigm. This acquired capability results in a massive performance leap across all benchmarks, demonstrating that the model receives effective supervision from our distilled data and that the ALR reasoning paradigm possesses strong generalization capabilities.

\noindent\textbf{The role of the Evidence-aware GRPO.} Although the SFT stage achieved significant success, we observe that the introduction of RL stage brought further, consistent performance improvements across all five benchmarks. This indicates that the ALR capabilities injected during SFT still had room for optimization and our Evidence-aware GRPO provided this crucial refinement.

We also provide extensive representative examples in Appendix F to corroborate our experimental conclusions.

\begin{table}[htbp]
 \centering
 \footnotesize
 \caption{Performance comparison of DocSeeker and the Baseline under two input conditions: Full-document input (Full-doc) and Evidence-only input (Evi-only). \textbf{Avg.}: Avg. input pages.}
 \label{tab:full_doc and gt_only}
 \renewcommand\arraystretch{1.00} 

 \begin{tabular}{lccll}
    \toprule
    \multirow{2}{*}{\textbf{Model}} & \multirow{2}{*}{Input} & \multirow{2}{*}{Avg.}&  \multicolumn{2}{c}{\textbf{MMLong.}}\\
    \cmidrule(lr){4-5}
    &  &   & Acc. & F1\\
    \midrule
    Baseline & Evi-only & 1.5  &  41.1 & 37.6 \\
    \midrule
    Baseline & Full-doc & 43.4 & 25.4\textcolor{red}{\scriptsize{(-15.7)}} & 20.8\textcolor{red}{\scriptsize{(-16.8)}} \\
    DocSeeker & Full-doc & 43.4 & 40.1\textcolor{red}{\scriptsize{(-1.0)}} & 38.4\textcolor{red}{\scriptsize{(+0.8)}} \\
    \bottomrule
 \end{tabular}
\end{table}

\subsection{Impact of Document Length on Performance}
\label{Sec:analysis}

To investigate the reasons for DocSeeker's superior performance, we conducted two experiments on MMLongBench-doc to analyze the impact of document length.

\textbf{Full-Document vs. Evidence-Only Input.}
We hypothesize that DocSeeker's advantage primarily stems from a robust evidence localization capability. To validate this, we compare performance under two distinct conditions: a controlled ``evidence-only'' input, which requires no localization and a ``full-document'' input, which demands localization amidst noise. The results, presented in Tab~\ref{tab:full_doc and gt_only}, show that  feeding the entire document led to a significant decline in accuracy, highlighting the limitations of current MLLMs in terms of document localization for long documents. In contrast, Docseeker's performance with full-document input closely matched that with evidence page input.

\begin{figure}[htbp]
  \centering
  \vspace{-10pt}
  \includegraphics[width=0.95\columnwidth]{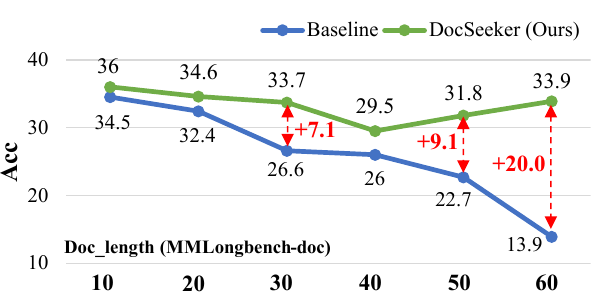}
  \caption{Performance comparison on different document length}
  \label{fig:doc_length}
\end{figure}

\textbf{Analysis of Document Length on Performance.} To further investigate the impact of document length in a controlled manner, we conducted an experiment on a subset of MMLongBench-doc, using only documents originally longer than 60 pages.  We fixed the question and ensured the evidence page was always included, while truncating the total context to different lengths. This allows us to isolate the total context length as the primary variable and precisely quantify its direct impact on performance. For this analysis, the "evidence-only" input condition serves as the theoretical upper bound for performance, representing perfect evidence localization. The results in Fig.~\ref{fig:doc_length} show that Baseline's performance dramatically degrades as the document length increases, with its accuracy plummeting from 34.5\% to 13.9\%. In contrast, DocSeeker exhibits remarkable robustness, as its performance remains largely stable.  This experiment re-confirms that our ALR paradigm enables the model to robustly localize evidence within long contexts, undeterred by the interference of irrelevant pages.

\begin{figure}[htbp]
  \makebox[\columnwidth][l]{%
    \hspace*{-0.8cm}
    \includegraphics[width=\dimexpr\columnwidth+1.2cm\relax]{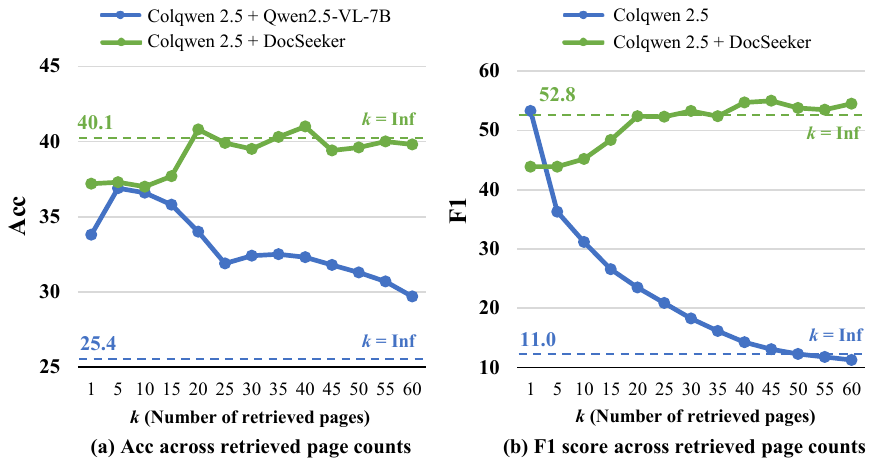}%
  }
  \caption{Performance comparison on MMLongBench-doc following integration with a retrievar Coilpail. \textbf{Acc} measures answer correctness, while \textbf{F1} quantifies evidence page retrieval. The k=Inf designation indicates the result for full-document input.}
  \label{fig:combine_with_retrieval}
\end{figure}

\subsection{Integration with RAG-based systems}
Sec.~\ref{Sec:analysis} has demonstrated the strong evidence localization capabilities of DocSeeker, enabling its performance on full input to closely approach that of ideal retrieval ground truth input. This core capability for precise localization within long contexts makes our model an ideal choice for integration with visual retriever (e.g., ColPail~\cite{faysse2024colpali}) in a RAG system to tackle multi-document or ultra-long document reasoning. This allows it to directly address a critical and well-known flaw in RAG: model performance collapses when the number of retrieved pages (K) is too small, which risks missing key evidence, or too large, which introduces excessive noise and causes the SNR to plummet. 

The experimental results clearly validate this phenomenon. As shown in Fig.~\ref{fig:combine_with_retrieval} (a), the performance of baseline collapses dramatically as K increases, especially for values greater than 5. This is because as the retriever increases K to improve Recall, its Precision and F1 sharply declines, as shown in Fig.~\ref{fig:combine_with_retrieval} (b). The baseline model, lacking the ability to perform localization in long and noisy contexts, is consequently overwhelmed by the noise. In contrast, DocSeeker effectively resists this noise interference. More importantly, integrating with the retrieval system further enhances DocSeeker's performance. This indicates a strong synergy: the retriever performs coarse-grained filtering, while our model efficiently conducts fine-grained reading and localization within the still-noisy retrieved results.

\begin{table}[t]
\centering
\caption{Ablation study on data types and size. The first group investigates different ALR CoT data configurations, while the second explores the effect of varying data size on performance.}
\label{tab:data-ablation}

\footnotesize
\setlength{\tabcolsep}{5pt}        
\renewcommand{\arraystretch}{0.95} 

\begin{tabularx}{0.8\linewidth}{l>{\hspace{16pt}}c Y>{\hspace{2pt}}Y}
\toprule
\multirow{2}{*}{\textbf{Data Config}} &
\multirow{2}{*}{\textbf{Size}} &
\multicolumn{2}{c}{\textbf{MMLong.}} \\
\cmidrule(lr){3-4}
& & Acc. & F1 \\
\midrule
\multicolumn{4}{l}{\emph{Data type ablations}} \\
\midrule
Baseline               & --   & 25.4 & 20.8 \\
Raw short-answer data  & 6.3k & 27.4 & 27.6 \\
Vanilla CoT            & 6.3k & 31.3 & 32.4 \\
\rowcolor{rowgray}
ALR CoT data              & 6.3k & \textbf{33.8} & \textbf{33.9} \\
-w/o Page id           & 6.3k & 30.4 & 31.1 \\
\midrule
\multicolumn{4}{l}{\emph{Data size ablations (ALR CoT data)}} \\
\midrule
20\% of ALR CoT data & 2.8k & 32.7 & 31.7 \\
40\% of ALR CoT data & 5.6k & 35.8 & 33.8 \\
60\% of ALR CoT data & 8.4k & 36.5 & 34.9 \\
80\% of ALR CoT data & 11k  & 38.2 & 36.5 \\
\rowcolor{rowgray}
All ALR CoT data     & 13k  & \textbf{38.6} & \textbf{36.9} \\
\bottomrule
\end{tabularx}
\end{table}

\subsection{Ablation Study}

We conducted comprehensive experiments on the MMLongBench-doc (hereafter referred to as MMLong.), chosen for its OOD characteristics regarding both data and document length distribution, to validate the effectiveness of the proposed method.

\noindent\textbf{Ablation on Reasoning Paradigm and Data Size.} To evaluate the impact of different reasoning paradigms and training data sizes on the model's performance, we conduct an ablation study with data types including Raw Short-Answer Data, Vanilla CoT Data (with unstructured reasoning and answers also distilled from the same teacher model Gemini-2.5-Flash), and our ALR CoT Data, trained for two epochs. The generation process for Vanilla CoT and ALR CoT Data is detailed in Appendix A. To ensure fair comparison, we select the intersection subset of the three datasets after correctness validation. Additionally, we also verify the impact of Page ID and data size in the ALR CoT Data. The corresponding In-domain evaluation results for different reasoning paradigms and data sizes are detailed in Appendix B.
The experimental results in Tab.~\ref{tab:data-ablation} indicate that: 1) Raw Short-Answer Data shows limited generalization, providing only marginal gains. 
2) Vanilla CoT Data offers modest improvements, while our ALR CoT data provides a more significant boost. 3) Larger-scale ALR CoT data consistently enhances performance. Additionally, removing Page ID reduces performance, highlighting its crucial role in providing page-level context and aiding information retrieval.

\begin{table}[htbp]
\centering
\vspace{-8pt}
\caption{Ablation study on training resolution strategies. “Fixed-Res” denotes fixed resolution for all pages, while EGRA adjusts page resolution during training based on evidence relevance. Tokens indicate the number of visual tokens used to encode an image.}
\label{tab:mm-longbench-res}
\setlength{\tabcolsep}{6pt}
\resizebox{\linewidth}{!}{
\begin{tabular}{llccc}
\toprule
\multirow{2}{*}{\textbf{Method}} & 
\multirow{2}{*}{\textbf{Strategy}} &
\multirow{2}{*}{\textbf{Tokens}} & 
\multicolumn{2}{c}{\textbf{MMLong.}} \\
\cmidrule(lr){4-5}
 & & & Acc. & F1 \\
\midrule
Baseline & - & - &25.4 & 20.8\\
\hline
\multirow{2}{*}{Fixed-Res} 
  & Full Low-Res. & 576 & 34.2 & 33.5 \\
  & Truncated & 1024 & 36.6 & 35.8\\
\hline
\multirow{3}{*}{\makecell[c]{EGRA}}
  & Full &1024 or 256 & \textbf{38.6} & \textbf{36.9} \\
  & -w/o Non-Evi. Hi-Res & 1024 or 256 & 35.5 & 33.8 \\
  & -w/o Low-Res & 1024 & 34.5 & 33.4 \\
\bottomrule

\end{tabular}
}
\end{table}

\noindent\textbf{Ablation on Resolution Allocation Strategy.} To investigate the effect of resolution allocation on model performance under a fixed token budget, we conduct an ablation study comparing our proposed EGRA strategies with other strategies. The Fixed Allocation strategy either employs higher resolution with truncated inputs or full-page inputs at lower resolution. EGRA (w/o Low-Res) refers to directly removing 70\% of none-evidence pages (instead of retaining low resolution), while EGRA (w/o Non-Evi. Hi-Res) refers to reducing the resolution of all none-evidence pages. The results in Tab.~\ref{tab:mm-longbench-res} show that: 1) Fixed Allocation settings underperform compared to EGRA. Because fixed resolution training presents a trade-off: higher resolution risks losing context, while lower resolution provides limited supervision, both leading to the loss of effective information during training; 2) Non-evidence pages, though not contributing to reasoning, retaining them at a low resolution is superior to discarding; 3) Selectively downsampling a random subset of non-evidence pages, rather than all of them, enhances model robustness by introducing high-resolution distractors, which also mitigates the resolution inconsistency between training and inference phases. EGRA effectively balances detail and context length, optimizing resolution and computational efficiency, leading to enhanced long-document performance and generalization.

\begin{table}[t]
\centering
\caption{Ablation on GRPO variants and reward weights. Vanilla GRPO optimizes answer correctness only; EviGRPO further includes localization-related rewards.}
\label{tab:grpo-ablation}

\footnotesize                     
\setlength{\tabcolsep}{5pt}       
\renewcommand{\arraystretch}{0.95}

\begin{tabular}{l c cc}
\toprule
\multirow{2}{*}{\textbf{Method}} &
\multicolumn{1}{c}{\textbf{Weight}} &
\multicolumn{2}{c}{\textbf{MMLong.}} \\
& {\textbf{$(\lambda_{1},\,\lambda_{2},\,\lambda_{3})$}} & Acc. & F1 \\
\midrule
SFT & -- & 38.6 & 36.9 \\
\midrule
Vanilla GRPO & $(0.1,\ 0.0,\ 0.9)$ & 38.7 & 36.3 \\
\midrule
        & $(0.1,\ 0.1,\ 0.8)$ & 39.4 & 37.8 \\
\rowcolor{rowgray}
EviGRPO & $(0.1,\ 0.3,\ 0.6)$ & \textbf{40.1} & \textbf{38.4} \\
        & $(0.1,\ 0.5,\ 0.4)$ & 38.9 & 37.0 \\
\bottomrule
\end{tabular}
\end{table}

\noindent\textbf{Ablation on RL Configurations.} To evaluate the effect of GRPO variants and reward weights on DocSeeker's performance, we conducted a series of ablation experiments. The experiment results presented in Tab.~\ref{tab:grpo-ablation} suggest that
The proposed Evidence Localization Reward effectively assists the Answer Reward, further improving the model’s performance and generalization in long-document understanding tasks.
Furthermore A large Evidence Localization weight causes the model to overly focus on evidence localization, leading to recognition errors, where incorrect answers may receive high reward; whereas a small weight may cause the model to ignore evidence localization signals.\\

\section{Conclusion}

In this work, we present DocSeeker, which adopts the novel ALR visual reasoning paradigm, a two-stage training framework and an effective resolution strategy to address the core challenges of low signal-to-noise ratio (SNR) and supervision scarcity in long-document understanding. DocSeeker learns fine-grained evidence localization and achieves robust long-document reasoning, even when trained on existing relatively short multi-page datasets. Experimental results demonstrate that our method achieves satisfactory performance and generalization on long document understanding. Furthermore, visual RAG integration experiments confirm that DocSeeker effectively alleviates the inherent limitations of visual RAG system, exhibiting remarkable robustness to retrieval noise and achieving strong synergy with the retriever. These results highlight DocSeeker as a promising foundation for building efficient and reliable RAG-based multimodal reasoning systems.

\section*{Acknowledgments}
This work was supported by NSFC (No. 62225603, No. 62576147, No. U25A20538, No. U25B2078).

{
    \small
    \bibliographystyle{unsrt}
    \bibliography{main}
}



\appendix
\twocolumn[
\begin{center}
\large \textbf{APPENDIX}
\end{center}
\vspace{0.5em}
]

\section{Details of Data Distillation}
\label{sec:Details of Data Distillation}
\subsection{Overview of the Distillation Pipeline}
\label{sec:distillation}

\begin{figure*}[t]
  \centering
  \includegraphics[width=\textwidth]{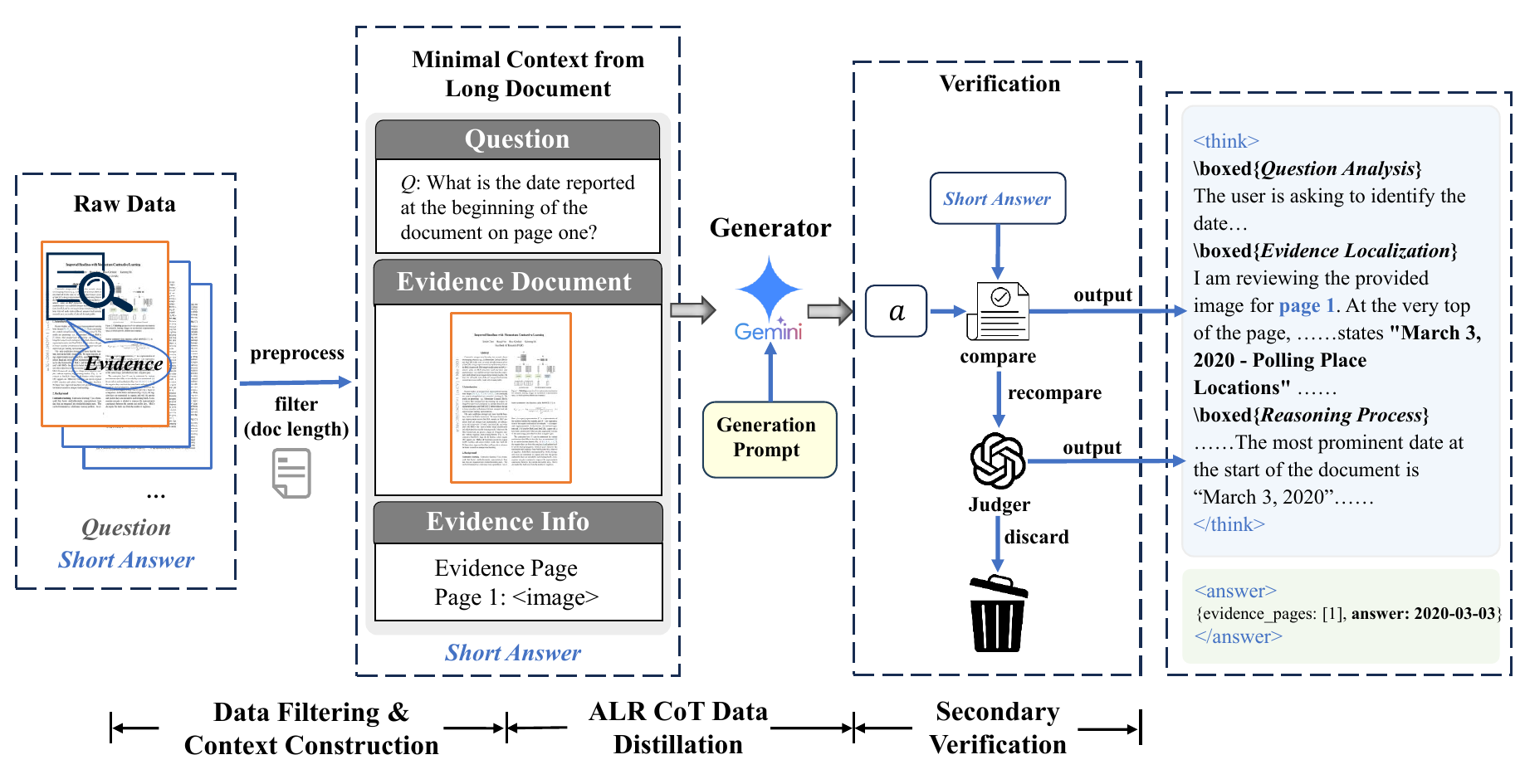}
  \caption{Data Distillation Pipeline.}
  \label{fig:overvew_of_distillation}
\end{figure*}

To address the issue of supervision scarcity in existing multi-page document VQA datasets, which typically provide only final short answers and evidence page indices without intermediate reasoning steps, we designed a rigorous data distillation pipeline. As illustrated in Figure~\ref{fig:overvew_of_distillation}, this pipeline transforms raw samples into high-quality structured ALR CoT data. The overall process consists of three key phases:

\textbf{Data Filtering and Context Construction.} We strictly select long-document samples from MP-DocVQA and DUDE, applying a pre-filtering step to discard documents that are overly short. Subsequently, we construct a "Minimal Context" for each sample, composed exclusively of the evidence pages, corresponding Page IDs, and the question. This strategy significantly reduces distillation costs while substantially enhancing the quality of generated data. Quantitative analysis on a subset of 1,000 long documents (avg. 17.6 pages) demonstrates that this approach improves the distillation success rate from 20.4\% (full-context input) to 67.3\%.

\textbf{ALR CoT Data Distillation.} We employ Gemini-2.5-Flash~\cite{comanici2025gemini}  as the teacher model due to its strong multimodal reasoning capabilities. By utilizing a specialized prompt, we instruct the teacher to act as an expert annotator. The teacher generates a structured response that explicitly includes Question Analysis, Evidence Localization (citing specific Page IDs), and a Reasoning Process, strictly following our proposed ALR paradigm.

\textbf{Secondary Verification.} To guarantee the correctness of the synthesized reasoning paths, we implement a robust two-stage verification mechanism. First, we apply an automated Exact Match (EM) filter to check if the generated final answer and evidence page IDs align perfectly with the ground truth. For samples that fail the strict EM check (e.g., due to paraphrasing), we employ GPT-4o as a semantic judge to validate the correctness of the answer . Only samples passing this verification are retained for the SFT stage.

\subsection{Prompt Templates}

This section outlines the specific prompt templates integral to our data distillation pipeline. Figure~\ref{fig:alr_cot_prompt} presents the ALR CoT Data Distillation Prompt, designed to extract high-quality structured reasoning from the teacher model. Furthermore, we provide the Vanilla CoT Data Distillation Prompt in Figure~\ref{fig:vanilla_cot_prompt} for ablation studies; unlike the ALR approach, this prompt encourages the model to generate free-form reasoning chains without enforcing the structured constraints of question analysis or explicit evidence localization. Finally, Figure~\ref{fig:data_filter} illustrates the LLM-based Judge Prompt, employed during secondary verification to salvage factually correct samples initially discarded due to formatting mismatches.

\begin{table*}[h]
\centering
\small
\caption{Ablation study on data types and training data size. The first group investigates different CoT data configurations, while the second explores the effect of varying data size on performance.}
\label{tab:data-ablation-all}

\setlength{\tabcolsep}{15pt}  

\begin{tabular}{lcccccc}
\toprule
\multirow{2}{*}{\textbf{Data Config}} & 
\multirow{2}{*}{\textbf{Size}} & 
\multicolumn{2}{c}{\textbf{MMLongbench.}} & 
{\textbf{DUDE}} \\
\cmidrule(lr){3-4} \cmidrule(lr){5-5} 
 &  &  Acc. & F1 & ANLS \\
\midrule
Baseline & - & 25.4 & 20.8 & 35.2 \\
\midrule
Raw short-answer data & 6.3k & 27.4 & 27.6 & 48.8 \\
Vanilla CoT data & 6.3k & 31.3 & 32.4 & 48.7 \\
ALR CoT data & 6.3k & \textbf{33.8} & \textbf{33.9} & 48.9 \\
-w/o Page id & 6.3k & 30.4 & 31.1 & 47.5 \\
\midrule
20\% of ALR CoT data & 2.8k & 32.7 & 31.7 & 46.5 \\
40\% of ALR CoT data & 5.6k & 35.8 & 33.8 & 52.3 \\
60\% of ALR CoT data & 8.4k & 36.5 & 34.9 & 54.5 \\
80\% of ALR CoT data & 11k & 38.2 & 36.5 & 55.9 \\
All ALR CoT data & 13k & \textbf{38.6} & \textbf{36.9} & \textbf{56.5} \\
\bottomrule
\end{tabular}
\end{table*}

\subsection{Failure Analysis}
\begin{figure*}[h]
  \centering
  \includegraphics[width=\textwidth]{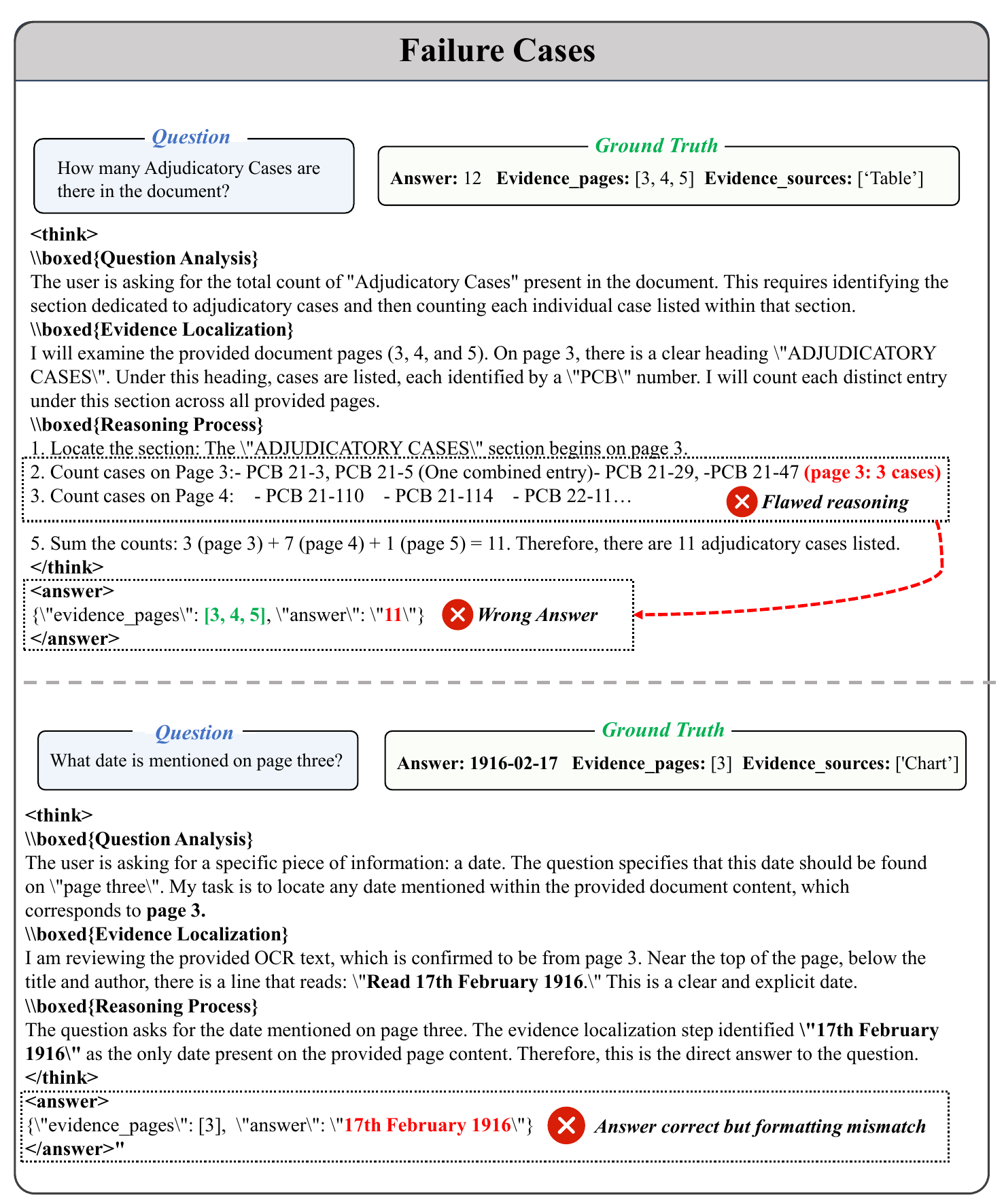}
  \caption{Failure cases in ALR CoT Data Distillation.}
  \label{fig:fail}
\end{figure*}

We visualize representative examples of samples that failed the initial automated Exact Match screening in Figure~\ref{fig:fail},  which fall into two distinct categories:

1) \textbf{Genuine Reasoning Failures.} These cases contain substantive quality defects. Despite attempting to follow the ALR paradigm, the model commits logical errors during the reasoning process, resulting in factually incorrect answers. These constitute actual noise and must be strictly discarded.

2) \textbf{Correct Reasoning with Formatting Mismatches.} These samples exhibit correct analysis, localization, and reasoning process.  The derived answers are semantically correct but fail the EM check solely due to  discrepancies in formatting, e.g. 17th February 1916 vs 1916-02-17. Such samples also represent high-quality supervision signals that should be preserved to maximize data utilization.

These observations underscore the necessity of the Secondary Verification mechanism introduced in Section~\ref{sec:distillation}, which is essential for accurately distinguishing valid semantic matches from genuine reasoning errors.

\begin{figure*}[p]
    \centering
    \begin{tcolorbox}[colframe=black, colback=white]
    \begin{center}
    \textbf{ALR CoT Data Distillation Prompt}
    \end{center}
    \textbf{Task Definition:} \\
    You are an AI assistant specialized in constructing high-quality question-answering datasets. Your core task is to generate a detailed reasoning process and a complete, natural answer based on the given question, the corresponding document evidence page, and the page number. \\
    
    \textbf{Important Rule:} \\
    All reasoning and the final answer must be strictly derived from the provided \textit{Document Page Content}, without using external knowledge or unsupported guesses. \\
    
    \textbf{Output Requirements:} \\
    Your output must strictly contain two sections, ``\texttt{<think>}'' and ``\texttt{<answer>}'' following the exact format below. \\
    
    \textbf{Expected Output Format:} 
\begin{verbatim}
Reasoning Process ("<think>" section)
This section must use "\boxed{}" to wrap the following three subhead
-ings to clearly present your reasoning steps:
<think>
\boxed{Question Analysis}
    Analyze the intent of the question in depth. Clearly identify what 
    entity, concept, quantity, or relationship the question is asking 
    about. Break complex questions down if needed.
\boxed{Evidence Localization}
    Identify which page contains the evidence, then locate the key 
    content within the provided Document Page Content.
\boxed{Reasoning Process}
    Derive the answer step by step using only localized evidence. 
    Explain how the conclusion is formed.
</think>
Final Answer ("<answer>" section)
This section must be a single JSON object containing:
<answer>
evidence_pages: A JSON list including the page numbers.
answer: A concise string giving the direct answer without introductory 
        phrases or full sentence structures.
</answer>
\end{verbatim}

\textbf{Example:} \\
    \textbf{Question:} mention the texts in brackets with bold fonts in paragraph 2 in page 5? \\
    \textbf{Document Page Number:} [5] 
\begin{verbatim}
<think>
\boxed{Question Analysis}
The user asks for bracketed text in paragraph 2 on page 5 that is 
also bolded.
\boxed{Evidence Localization}
Page 5 contains bracketed text but no bolded text.
\boxed{Reasoning Process}
No text meets all required conditions; therefore the question cannot 
be answered.
</think>
<answer>
{"evidence_pages": [5], "answer": "Not answerable"}
</answer>
\end{verbatim}
    \end{tcolorbox}
    \caption{ALR CoT Data Distillation Prompt Templete.}
    \label{fig:alr_cot_prompt}
\end{figure*}

\begin{figure*}[h]
    \centering
    \begin{tcolorbox}[colframe=black, colback=white]
    \begin{center}
    \textbf{Vanilla CoT Data Distillation Prompt}
    \end{center}
    
    \textbf{Task Definition:} \\
    You are an AI assistant specialized in constructing high-quality question-answering datasets. Your core task is to generate a detailed reasoning process and a complete, natural answer based on the given question, the corresponding document evidence page, and the page number.\\[4pt]
    
    \textbf{Requirements:} \\
    \begin{itemize}
        \item All your reasoning and final answer must be strictly derived from the provided ``Document Page Content''.
        \item You are not allowed to use any external knowledge or make unsupported guesses.
        \item Your output must strictly follow the format below with two required sections: ``\texttt{<think>}'' and ``\texttt{<answer>}``.\\[4pt]
    \end{itemize}

    \textbf{Expected Output:} \\
\verb|<think>| \\
Let's think step by step. \\
... \\
\verb|</think>| \\[4pt]
\verb|<answer>| \\
A string that provides a direct and concise answer to the question without any introductory phrases or full sentence structures. \\
\verb|</answer>|
    \end{tcolorbox}
    
    \caption{Vanilla CoT Data Distillation Prompt Templete.}
    \label{fig:vanilla_cot_prompt}
\end{figure*}

\section{Additional Ablation Study Results}
\label{sec:Additional_Ablation_Study_ Results}
To delve deeper into the impact of distinct reasoning paradigms and data strategies on Generalization Capabilities, we supplemented the In-domain (DUDE~\cite{van2023document}) evaluation in Table~\ref{tab:data-ablation-all}. SFT on Raw Short-Answer data yields In-domain gains comparable to other paradigms, yet improvement on OOD tasks remains negligible. This suggests the model primarily engages in Rote Memorization rather than mastering the general reasoning skills required for unseen long documents. In contrast, increasing the scale of ALR CoT data not only progressively enhances In-domain performance but, critically, achieves a synchronous leap in OOD tasks. This demonstrates that the structured "Analysis-Localization-Reasoning" paradigm instills transferable reasoning capabilities, enabling the model to transcend specific data distributions and actively adapt to unfamiliar long-document structures, thereby achieving true generalization.

\section{Efficiency and Accuracy Trade-offs}
Although the proposed ALR paradigm enhances the model's capability in long-document understanding, it inevitably introduces modest additional computational latency. To investigate the trade-off between inference latency and accuracy, we conducted experiments on MMLongBench-Doc, as shown in Table \ref{tab:token_latency_acc}. Although the ALR paradigm nearly doubles output tokens, the end-to-end latency increases only modestly from 19s to 25s. This is because the primary computational bottleneck in long-document processing is the visual pre-fill stage rather than the text decoding stage. Ultimately, DocSeeker trades a 31.6\% latency increase for a 14.7\% absolute accuracy gain and enhanced interpretability.

\begin{table}[h]
\centering
\caption{Analysis of inference latency and accuracy trade-off on MMLongBench-Doc}
\begin{tabular}{lccc}
\toprule
&
Token &
Latency &
Acc \\
\midrule
Baseline  & 202 & 19s & 25.4 \\
DocSeeker & 401 & 25s & 40.1 \\
\bottomrule
\end{tabular}
\label{tab:token_latency_acc}
\end{table}

\begin{table}[h]
\centering
\caption{Stage-wise error breakdown on MMLongBench-Doc.}
\setlength{\tabcolsep}{10pt}
\renewcommand{\arraystretch}{0.85}
\begin{tabular}{lccc}
\toprule
  & \textbf{Acc$\ge 0.5$} & \textbf{Acc$=0$} & \textbf{Total} \\
\midrule
\textbf{Recall$=1$} & 300  & 288  & 588 \\
\textbf{Recall$<1$} & 121  & 373  & 494 \\
\textbf{Total} & 421 & 661 & 1082 \\
\bottomrule
\end{tabular}

\vspace{-1.2em}
\label{tab:error_breakdown}
\end{table}

\begin{figure*}[h]
    \centering
    \begin{tcolorbox}[colframe=black, colback=white]
    \begin{center}
    \textbf{Judge Prompt}
    \end{center}
    
    \textbf{Task Definition:} \\
    You are an expert AI assistant for data validation and correction. Your core task is to compare a model-generated response with a ground-truth answer and determine the final, most accurate output based on a set of rules. \\
    
    \textbf{Inputs:} \\
    You will be given three pieces of information: \textit{question}, \textit{response}, and \textit{answer}. \\
    
    \textbf{Requirements:} \\
    Your output should be a single, clean string representing the corrected answer, or the word `Error' if applicable. \\
    
    \textbf{Rules:}
    \begin{enumerate}
        \item \textbf{Formatting Mismatch:} If the response and answer contain the same core information but differ only in formatting (e.g., date separators like / vs. -, number separators like , vs. ., or case sensitivity), you must output the version from the ground-truth answer.
        \item \textbf{Overly Detailed Response:} If the response is factually correct and contains all the information from the answer, but includes additional conversational phrases or details (e.g., ``The answer is X'' vs. ``X''), you should conservatively output the more concise ground-truth answer.
        \item \textbf{Overly Simplified Response:} If the response is correct but misses key details that are present in the answer, you must output the more complete ground-truth answer.
        \item \textbf{Correct Response, Flawed Answer:} If you determine that the response is factually or grammatically more correct than the ground-truth answer (e.g., the answer has a clear typo or factual error that the response corrects), you must output the model-generated response to ensure data quality.
        \item \textbf{Factually Incorrect/Contradictory:} If the response and answer are factually contradictory and irreconcilable (e.g., different names, numbers, dates), or if one provides a factual answer while the other states it's not answerable, you must output the exact string \texttt{Error}.
    \end{enumerate}
    
    \textbf{Output Format:} \\
    \texttt{output:}\\
    \texttt{<answer>}\\[4pt]
    
    Please read the following example and complete the output in the specified format based on the given information.\\

    \textbf{Example: }

    \textbf{Question:} \\
    WHAT IS THE APPROVED DATE FOR RELEASE? \\[6pt]
    \textbf{Response:} \\
    2001/07/27 \\[6pt]
    \textbf{Answer:} \\
    2001-07-27 \\[6pt]
    \textbf{Output:} \\
    2001-07-27
    \end{tcolorbox}
    \caption{LLM-based Judge Prompt Templete in Secondary Verification.}
    \label{fig:data_filter}
\end{figure*}

\section{Stage-wise Error Breakdown}
To systematically decompose the errors encountered  in long-document understanding, we perform a stage-wise error breakdown on the samples from MMLongBench-Doc. As detailed in Table \ref{tab:error_breakdown}, we categorize these samples based on the correlation between evidence recall (Recall) and answer accuracy (Acc), which reveals three primary failure modes:

(i) \textbf{Reasoning failures.} The model successfully retrieves all required evidence pages ($Recall=1$) but fails to provide the correct final deduction

(ii) \textbf{Localization failures.} The model fails to localize the evidence pages, leading to a lack of sufficient context for the model to perform downstream reasoning.

(iii) \textbf{Exceptions.} The model provides correct answers despite incomplete localization ($Recall < 1$) by leveraging its internal global context awareness.

\begin{table*}[h]
\centering
\caption{Detailed Configurations of Compared Methods on Long Document Understanding.}
\label{tab:sp_results_final_v16}
\fontsize{9}{11}\selectfont
\setlength\arrayrulewidth{0.6pt}
\setlength{\tabcolsep}{2pt}
\resizebox{1\textwidth}{!}{
\begin{tabular}{l l c c c c c c c}
\hline
\multirow{2}{*}{\textbf{Method}} & \multirow{2}{*}{\textbf{Model}} & \multirow{2}{*}{\textbf{OCR}} & \multicolumn{2}{c}{\textbf{Retrieval Config}} & \multirow{2}{*}{\textbf{Backbone}} & \multicolumn{3}{c}{\textbf{Trained on Dataset}} \\
\cmidrule(lr){4-5} \cmidrule(l){7-9} 
 & & & \textbf{Retriever} & \textbf{Top-k} & & \textbf{DUDE} & \textbf{MPDocVQA} & \textbf{SlideVQA} \\
\hline

\multirow{5}{*}{\textbf{RAG}} 
 & CREAM~\cite{zhang2024cream} & $\checkmark$ & bge-large~\cite{bge_embedding} & 3 & Pix2Struct/LLaMa2 & $\checkmark$ & $\checkmark$ & $\times$ \\
 & M3DocRAG~\cite{cho2024m3docrag} & $\times$ & Colpali~\cite{faysse2024colpali} & 4 & Qwen2-VL & $\times$ & $\times$ & $\times$ \\
 & Vis-RAG~\cite{yu2024visrag} & $\times$ & VisRAG-Ret~\cite{yu2024visrag} & 3 & MiniCPM-V 2.6 & $\times$ & $\times$ & $\times$ \\
 & SV-RAG~\cite{chen2024sv} & $\times$ & SV-RAG-InternVL2~\cite{chen2024sv} & 5 & SV-RAG-InternVL2 & $\times$ & $\times$ & $\checkmark$ \\
 & VDocRAG~\cite{tanaka2025vdocrag} & $\times$ & VDocRetriever~\cite{tanaka2025vdocrag} & 3 & VDocGenerator & $\checkmark$ & $\times$ & $\times$ \\
\midrule 

\multirow{5}{*}{\textbf{End-to-End}} 
 & HiVT5~\cite{tito2023hierarchical} & $\checkmark$ & - & - & DiT/T5 & $\times$ & $\checkmark$ & $\times$ \\
 & mPLUG-DocOwl2~\cite{hu2025mplug2} & $\times$ & - & - & ViT/LLaMa+MAM & $\checkmark$ & $\checkmark$ & $\times$ \\
 & Docopilot~\cite{mathur2024docpilot} & $\times$ & - & - & InternVL2 & $\checkmark$ & $\checkmark$ & $\times$ \\
 & DocVLM~\cite{nacson2025docvlm} & $\checkmark$ & - & - & Qwen2-VL & $\times$ & $\times$ & $\times$ \\
 & InternVL3~\cite{zhu2025internvl3} & $\times$ & - & - & InternViT/Qwen2.5 & $\times$ & $\times$ & $\times$ \\
\hline \hline
\end{tabular}
}
\end{table*}

\section{Details of Compared Methods}
\label{sec:E_Details_of_Compared_Methods}
Table~\ref{tab:sp_results_final_v16} details the specific evaluation configurations for the compared models across five benchmarks. For all baselines, we strictly adhered to the officially recommended or optimal settings to ensure a fair comparison.

\section{Case Studies}
\label{sec:case_study}
In this section, we present qualitative visualizations of representative examples to demonstrate the superior capabilities of DocSeeker in handling complex long-document understanding tasks.

\subsection{Comparative Analysis of Reasoning Paradigms}
\begin{figure*}[p]
  \centering
  \includegraphics[width=0.95\textwidth]{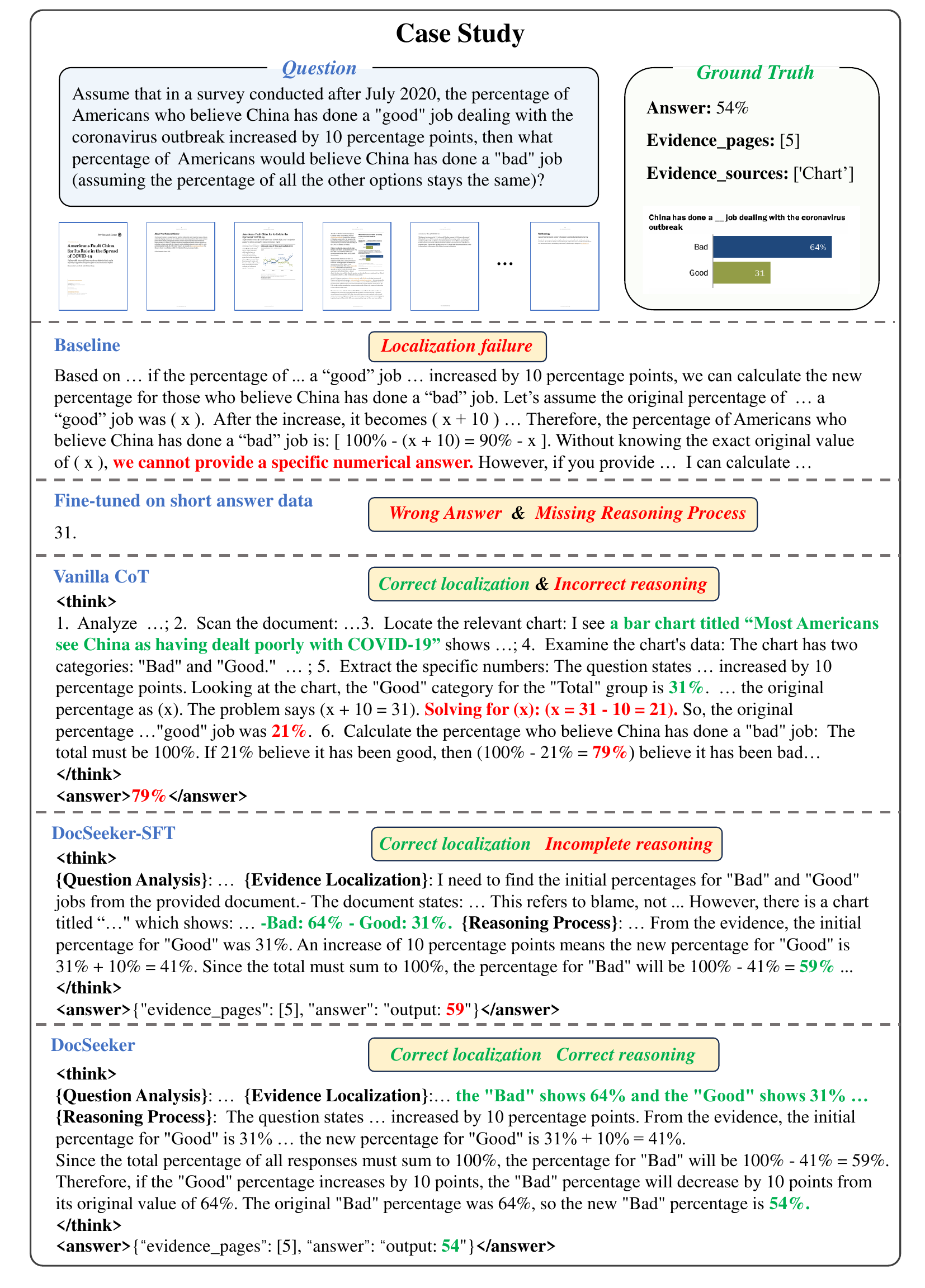}
  \caption{Qualitative Comparison of Reasoning Paradigms.}
  \label{fig:case2}
\end{figure*}

Figure~\ref{fig:case2} presents a qualitative comparison of DocSeeker against alternative reasoning paradigms. Experimental results indicate that the traditional Baseline model, constrained by limited long-context processing capabilities, exhibits complete localization failure, whereas the Short-Answer model yields "black-box" predictions devoid of logical support.  Vanilla CoT successfully localizes evidence, it suffers from "reasoning drift" during the unstructured generation process, resulting in deviations within the logical chain.  In contrast, DocSeeker leverages the structured ALR paradigm to strictly decouple evidence acquisition from logical deduction. This constraint mechanism effectively mitigates reasoning instability and ensures the execution of rigorous inference steps based on precise evidence, thereby demonstrating the significant robustness of our paradigm in complex long-document understanding.

\begin{table*}[h]
\centering
\caption{Comparison of method architectures and attributes, where "OCR" and "Retriever" denote dependency on external modules, "Training-free" indicates the absence of additional fine-tuning on multi-page documents, "Page-level Reasoning" refers to processing at page granularity, and "Evidence Localization" marks the capability for explicit source grounding.}
\label{tab:sp_results_modified}
\fontsize{9}{11}\selectfont
\setlength\arrayrulewidth{0.6pt}
\setlength{\tabcolsep}{2pt}
\resizebox{1\textwidth}{!}{
\begin{tabular}{l c c c c c c} 
\hline
\textbf{Method} & \textbf{OCR} & \textbf{Retriever} & \textbf{Backbone} &
\makecell[c]{\textbf{Training-}\\\textbf{free}} &
\makecell[c]{\textbf{Page-level}\\\textbf{Reasoning}} &
\makecell[c]{\textbf{Evidence}\\\textbf{Localization}} \\
\hline

\multicolumn{7}{l}{\textcolor{gray}{\fontsize{7}{8}\textit{OCR-based}}} \\
Longformer~\cite{beltagy2020longformer} & $\checkmark$ & $\times$ & RoBERTa & $\checkmark$ & $\times$ & $\times$ \\
DocLLM~\cite{wang2023docllm} & $\checkmark$ & $\times$ & Falcon/LLaMa2 & $\checkmark$ & $\times$ & $\times$ \\
LayTokenLLM~\cite{zhu2025simple} & $\checkmark$ & $\times$ & LLaMa3/LLaMa2/Qwen1.5 & $\times$ & $\times$ & $\times$ \\ 
\hline

\multicolumn{7}{l}{\textcolor{gray}{\fontsize{7}{8}\textit{End-to-End}}} 
\\
HiVT5~\cite{tito2023hierarchical} & $\checkmark$ & $\times$ & DiT/T5 & $\times$ & $\times$ & $\times$ \\
mPLUG-DocOwl2~\cite{hu2025mplug2} & $\times$ & $\times$ & ViT/LLaMa+MAM & $\times$ & $\checkmark$ & $\times$ \\
Docopilot~\cite{mathur2024docpilot} & $\times$ & $\times$ & InternVL2 & $\times$ & $\checkmark$ & $\times$ \\
DocVLM~\cite{nacson2025docvlm} & $\checkmark$ & $\times$ & Qwen2-VL & $\checkmark$ & $\times$ & $\times$ \\
InternVL3~\cite{zhu2025internvl3} & $\times$ & $\times$ & InternViT/Qwen2.5 & $\checkmark$ & $\checkmark$ & $\times$ \\
Qwen2.5VL~\cite{bai2025qwen25}  & $\times$ & $\times$ & QwenViT/Qwen2.5 & $\checkmark$ & $\checkmark$ & $\times$ \\
\hline

\multicolumn{7}{l}{\textcolor{gray}{\fontsize{7}{8}\textit{RAG}}} \\
PDF-WuKong~\cite{xie2024wukong} & $\checkmark$ & bge-m3~\cite{bge_embedding} & XComposer2-4KHD & $\times$ & $\times$ & $\checkmark$ \\
CREAM~\cite{zhang2024cream} & $\checkmark$ & bge-large~\cite{bge_embedding} & Pix2Struct/LLaMa2 & $\times$ & $\times$ & $\checkmark$ \\
M3DocRAG~\cite{cho2024m3docrag} & $\times$ & Colpali~\cite{faysse2024colpali} & Qwen2-VL & $\checkmark$ & $\checkmark$ & $\checkmark$ \\
Vis-RAG~\cite{yu2024visrag} & $\times$ & VisRAG-Ret~\cite{yu2024visrag} & MiniCPM-V 2.6 & $\checkmark$ & $\checkmark$ & $\checkmark$ \\
SV-RAG~\cite{chen2024sv} & $\times$ & SV-RAG-InternVL2~\cite{chen2024sv} & SV-RAG-InternVL2 & $\times$ & $\checkmark$ & $\checkmark$ \\
VDocRAG~\cite{tanaka2025vdocrag} & $\times$ & VDocRetriever~\cite{tanaka2025vdocrag} & VDocGenerator & $\times$ & $\checkmark$ & $\checkmark$ \\

\hline \hline 
\end{tabular}
}
\end{table*}

\begin{figure*}[t]
  \centering
  \includegraphics[width=\textwidth]{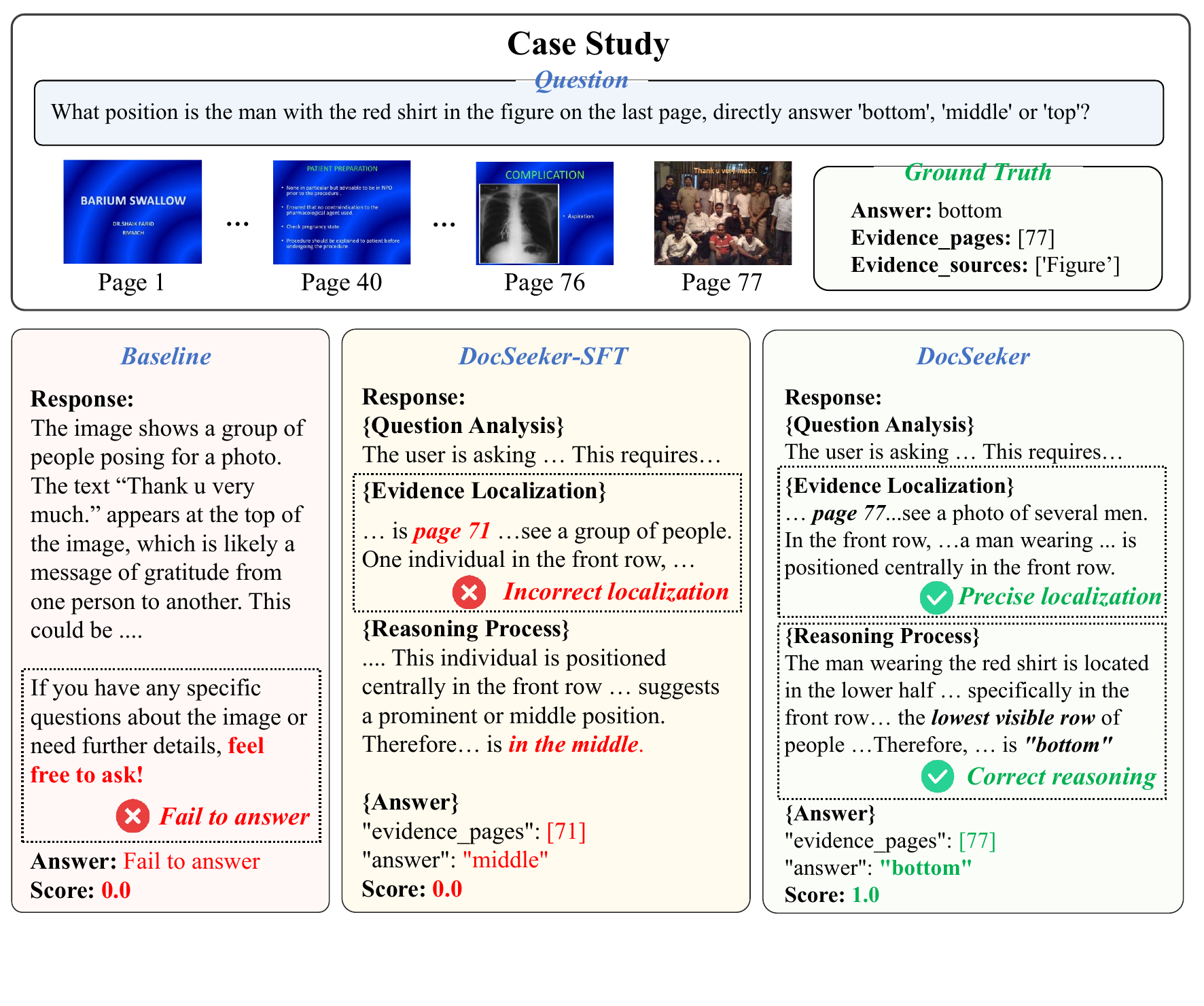}
  \caption{Performance Comparison across Different Training Stages.}
  \label{fig:case1}
\end{figure*}

\subsection{Efficacy of the Two-Stage Training Framework}
Figure~\ref{fig:case1} illustrates the evolution of model capabilities within the two-stage training framework. After SFT on ALR CoT data, DocSeeker-SFT successfully acquires the structural form of the ALR paradigm, strictly adhering to the "Analysis-Localization-Reasoning" workflow; however, it exhibits unstable grounding capabilities, evidenced by a incorrect page that results in an erroneous conclusion. This indicates that while SFT facilitates the imitation of reasoning patterns, it is insufficient for ensuring factual precision. The subsequent introduction of Evidence-aware GRPO effectively rectifies this limitation. Guided by localization-specific reward signals, the model achieves precise evidence grounding while retaining the structured reasoning path, demonstrating that the reinforcement learning stage is critical for optimizing localization accuracy and enhancing overall performance in long-document understanding.

\section{Related Work}
To provide a more intuitive comparison of distinct technical approaches in the field of long-document understanding, we summarize related studies in Table~\ref{tab:sp_results_modified}. These methods can be primarily categorized into the following three paradigms: 

1) OCR-based Methods: These approaches,such as LayTokenLLM~\cite{zhu2025simple} and DocLLM~\cite{wang2023docllm}, rely on external OCR engines to extract textual  information from documents, which then serves as input for the model. With the continuous advancement of OCR~\cite{liu2025multi,yin2025mstar}, these methods are becoming increasingly accurate and efficient. 

2) RAG-based Methods. To address the limitations of MLLMs in processing ultra-long documents, these method-

\newpage
\noindent
s, such as Vis-RAG~\cite{yu2024visrag} and VDocRAG~\cite{tanaka2025vdocrag}, introduce
 a retriever to select the Top-$k$ most relevant pages for subsequent reasoning. 

3) End-to-End Methods. These approaches, such as mPLUG-DocOwl2~\cite{hu2025mplug2} and InternVL3~\cite{zhu2025internvl3}, directly encode document images into visual tokens for processing, thereby preserving complete visual features.

\end{document}